# Real-Time Statistical Speech Translation

Krzysztof Wołk, Krzysztof Marasek

Department of Multimedia
Polish Japanese Institute of Information Technology, Koszykowa 86, 02-008 Warsaw
kwolk@pjwstk.edu.pl

**Abstract.** This research investigates the Statistical Machine Translation approaches to translate speech in real time automatically. Such systems can be used in a pipeline with speech recognition and synthesis software in order to produce a real-time voice communication system between foreigners. We obtained three main data sets from spoken proceedings that represent three different types of human speech. TED, Europarl, and OPUS parallel text corpora were used as the basis for training of language models, for developmental tuning and testing of the translation system. We also conducted experiments involving part of speech tagging, compound splitting, linear language model interpolation, TrueCasing and morphosyntactic analysis. We evaluated the effects of variety of data preparations on the translation results using the BLEU, NIST, METEOR and TER metrics and tried to give answer which metric is most suitable for PL-EN language pair.

**Keywords**: Machine translation, Speech translation, Machine learning, NLP, Knowledge-free learning.

## 1 Introduction

Automated translation of Polish-English speech, as compared to the other languages, is a formidable challenge. It is complicated declension, leading to a very wide range of vocabulary, its grammatical components and rules, cases, gender forms (7 and 15, respectively) for nouns and adjectives that drive its complexity. This complexity impacts both the Polish language data and data structures necessary for the Statistical Machine Translation (SMT).

Polish and English are very different in their syntax. Since English lacks declension endings, the way in which words are ordered in an English sentence is very significant for their meaning. English sentences follow a Subject-Verb-Object (SVO) pattern. The syntactic order often completely determines the meaning of a sentence.

On the other hand, syntactic order in Polish does not significantly influence the meaning of a sentence. It does not require any specific word order. For example, the English sentence "I went to cinema with my wife." is equivalent in Polish to many sentences like "Byłem w kinie z moją żoną.", "Z moją żoną byłem w kinie.", "W kinie byłem z moją żoną.", "Z moją żoną w kinie byłem.", "Byłem z moją żoną w kinie.", "W kinie z moją żoną byłem.".

Translation is greatly complicated by these differences in syntactic order between the two languages. This is particularly arduous when no additional lexical data is available and a phrase model [1] is used, which is often the case in the SMT systems.

The optimal input data for the SMT systems should have restricted subject domains like medical texts, historical books, European Parliament proceeding etc. It is very difficult for the SMT system to perform well using diverse domains. There are few Polish language corpora available to be used as input to the SMT systems. As a result, good sets of parallel Polish-English data have limited availability.

The baseline datasets used for this research were: Trans-language English Database (TED) [2], the Open Parallel Corpus[1] (OPUS) of movie subtitles, and the European Parliament (Europarl)[2] proceedings texts. The choice was not random. The TED represents lectures with no specific domain, OpenSubtitles are also not limited to any domain but they are a great example of casual human dialogs, that consist of many short sentences (easier to be translated by the SMT system), Europarl is an example of good quality in-domain data.

## 2   Data Preparation

There are approximately 2 million un-tokenized Polish words contained in the TED talks, 91 million in OpenSubtitles corpora and 15 million in Europarl. Preprocessing of this training information was both automatic and manual. Many errors were found in the data. Because of errors, dictionary size was increased and spelling errors degraded statistics. We extracted a Polish dictionary [3] containing 2,532,904 distinct words. Then, we created a dictionary containing 92,135 unique words from TED. Intersection of TED's dictionary with Polish dictionary, led to a new dictionary of 58,393 words. So, 33,742 Polish words that included spelling errors or named entities were found in TED. Very similar situation occurs in OpenSubtitles data but in the other hand, Europarl did not include many spelling errors, but it contained a lot of names and foreign words. After dealing with problems, final Polish-English TED, OpenSubtitles, Europarl corpora contained 134,678; 17,040,034; 632,565 lines respectively.

First, we used perplexity metrics to determine the quality of the available data. We used some data from the OPUS and some from the Europarl v7project. The rest was collected manually using web crawlers[3]. We created: PL–EN dictionary (Dict), newest TED Talks (TEDDL), e-books, Euro News Data, proceedings of UK Lords, subtitles for movies and TV series, Polish parliament and senate proceedings.

Table 1 provides the perplexity [4] metrics for our data. This shows, the perplexity values with Kneser-Ney smoothing [5] for Polish (PL) and for English (EN). Parallel data was noted in BI column. We used the MITLM [5] toolkit for that evaluation. The development data was used as the evaluation set for tuning. We randomly selected text from each corpora for development and testing, 1000 lines for each purpose.

---

[1] www.opus.lingfil.uu.se
[2] www.statmt.org/europarl
[3] www.korpusy.s16874487.onlinehome-server.info

These lines were deleted from the corpora for more reliable evaluation. The perplexity of the data was later analyzed.

**Table 1.** Data Perplexities

|  | TED | | EUROPARL | | OpenSubtitles | | Vocabulary Count | | |
| --- | --- | --- | --- | --- | --- | --- | --- | --- | --- |
| *Data* | *EN* | *PL* | *EN* | *PL* | *EN* | *PL* | *EN* | *PL* | *BI* |
| Baseline | 223 | 1,153 | 29 | 30 | 32 | 33 | 11,923 | 24,957 | + |
| Btec | 580 | 1,377 | 756 | 1,951 | 264 | 620 | 528,712 | 663,083 | + |
| Ebooks | 417 | 2,173 | 726 | 3,546 | 409 | 688 | 17,121 | 34,859 | - |
| ECB | 889 | 2,499 | 436 | 1,796 | 859 | 1,108 | 30,204 | 56,912 | + |
| EMEA | 1,848 | 3,654 | 1,773 | 4,044 | 1,728 | 1,233 | 167,811 | 228,142 | + |
| EUB | 950 | 3,266 | 219 | 874 | 1,276 | 2,120 | 34,813 | 45,063 | + |
| EUNEWS | 435 | 1,467 | 410 | 1,667 | 435 | 723 | 287,096 | - | + |
| GIGA | 610 | - | 415 | - | 822 | - | 343,468 | 480,868 | - |
| Other | 415 | 3,001 | 469 | 1,640 | 352 | 576 | 13,576 | 24,342 | + |
| KDE4 | 499 | 1,277 | 931 | 3,116 | 930 | 1,179 | 34,442 | 62,760 | + |
| News | 919 | 2,370 | 579 | - | 419 | - | 279,039 | - | - |
| NewsC | 377 | - | 595 | - | 769 | - | 62,937 | - | - |
| OpenSub | 465 | - | 1,035 | 6,087 | 695 | 1,317 | 47,015 | 58,447 | + |
| Dict | 594 | 2,023 | 8,729 | 59,471 | 2,066 | 1,676 | 47,662 | 113,726 | + |
| TEDDL | 8,824 | 40,447 | 539 | 1,925 | 327 | 622 | 39,214 | 39,214 | + |
| UK Lords | 644 | - | 401 | - | 721 | - | 215,106 | - | - |
| UN Texts | 714 | - | 321 | - | 892 | - | 175,007 | - | - |
| IPI | - | 2,675 | - | 774 | - | 1,212 | - | 245,898 | - |
| Lodz | - | 1,876 | - | 1,381 | - | 720 | - | 71,673 | - |
| Senat | - | 1,744 | - | 1,307 | - | 868 | - | 58,630 | - |
| Subtitles | - | 2,751 | - | 4,874 | - | 5,894 | - | 235,344 | - |
| TED TST | - | - | - | - | - | - | 2,861 | 4,023 | + |
| EU TST | - | - | - | - | - | - | 3,795 | 5,533 | + |
| OP TST | - | - | - | - | - | - | 1,601 | 2,030 | + |

EMEA designates texts from the European Medicines Agency. KDE4 is the localization file of user GUI. ECB is the European Central Bank corpus. OpenSubtitles are movie and TV series subtitles. EUNEWS is a web crawl of the euronews.com web page. EUBOOKSHOP comes from the bookshop.europa.eu website. BTEC is a phrasebook corpora, GIGA shortcut stands for a comprehensive archive of newswire text data that has been acquired by Linguistic Data Consoctium[1]. News and News Commentary data were obtained from WMT2012[2]. IPI is a large, morphosyntactically annotated, publicly available corpus of Polish[3]. SENAT stands for proceedings of the Polish Senate. Lastly, TEDDL is additional, TED data. Lastly we represent vocabulary sizes on each of data sets (TST suffix).

As can be seen in Tables 1 every additional data is much worse than the files provided in the baseline system, especially in case of Polish data. Due to differences in the languages and additoinal data, the text contained disproportionate vocabularies of 92,135 Polish and 41,163 English words as an example of TED data. Bojar [6] describes the use of word stems to address this problem. Convertion need was addressed by using the Wroclaw NLP tools [1]. The tools enabled us to use morphosyntactic analysis in our SMT system. This process also included tokenization, shallow parsing, and generation of feature vectors. The MACA framework was used to relate to the variety of morphological data, and the WCRFT [7] framework was used to produce combined conditional random fields. After it 40,346 stems remained in the PL vocabulary. This greatly reduced the disparity between the EN-PL lexicon size.

---

[1] www.catalog.ldc.upenn.edu/LDC2011T07
[2] www.statmt.org/wmt12
[3] www.korpus.pl

## 3 Factored Training

For training we used the open source Moses toolkit[2] that provides a range of training, tuning, and other SMT tools. The toolkit enables to use efficient data formats, large variety of linguistic factors, and confusion network decoding.

Phrase-based translation models, used in many SMT systems, are unable to leverage many sources of rich linguistic information (e.g. morphology, syntax, and semantics) useful in the SMT. Factored translation models try to make the use of additional information and more general representations (e.g. lemmas vice surface forms) that can be interrelated.

The use of a factored model [8] affects the preparation of the training data, because it requires annotation of the data with regard to the additional factors. The Moses Experiment Management System(EMS)[3] supports the use of factored models and their training. We used the Moses's Parts of Speech (POS) tagger, Compound Splitter, and Truecasing tools to create additional linguistic information for our factored systems. The POS processing utilized the MXPOST tool [9].

The Polish language allows compounding, i.e. generation of new, long words by joining together other words. Final, longer compound word is known as a periphrase. Periphrases present an additional challenge to SMT systems. We used the EMS's compound splitting tool [10] to split the data compounds into word stems by comparing the geometric mean of the steams frequency to the compound word frequency.

We also used the TrueCaser tool from the Moses toolkit to convert the input words to upper case or lower case, as appropriate to improve SMT output quality. Using truecasing should improve the quality of text and enable the use of uncased and poorly cased text as corpora for SMT.

## 4 Evaluation Methods

To obtain quality measurements on the translations produced by various SMT approaches, metrics were selected to compare the SMT translations to high quality human translations. We selected the Bilingual Evaluation Understudy (BLEU), U.S. National Institute of Standards & Technology (NIST) metric, Metric for Evaluation of Translation with Explicit Ordering (METEOR), and Translation Error Rate (TER) for our research.

---

[1] www.nlp.pwr.wroc.pl
[2] www.statmt.org/moses
[3] statmt.org/moses/?n=FactoredTraining.EMS

According to Axelrod, BLEU [11] uses textual phrases of varying length to match SMT and reference translations. Scoring of this metric is determined by the weighted averages of those matches.

To encourage infrequently used word translation, the NIST [11] metric scores the translation of such words higher and uses the arithmetic mean of the *n*-gram matches. Smaller differences in phrase length incur a smaller brevity penalty. This metric has shown advantages over the BLEU metric.

The METEOR [11] metric also changes the brevity penalty used by BLEU, uses the arithmetic mean like NIST, and considers matches in word order through examination of higher order *n*-grams. These changes increase score based on recall. It also considers best matches against multiple reference translations when evaluating the SMT output.

TER [11] compares the SMT and reference translations to determine the minimum number of edits a human would need to make for the translations to be equivalent in both fluency and semantics. The closest match to a reference translation is used in this metric. There are several types of edits considered: word deletion, word insertion, word order, word substitution, and phrase order.

## 5 Experimentation

We conducted experiments on phrase-based system as well as factored system enriched with POS tags. The use of compound splitting and true casing was optionary. Some language models based on perplexity measure were chosen and linearly interpolated [3].

We used the EMS to conduct the experiments. In addition, we implemented 5-gram language model training using the SRI Language Modeling Toolkit [4], together with interpolated Kneser-Key discounting. MGIZA++ tool [12], was used to align texts at the word and phrase level and the symmetrization method was set to grow-diag-final-and [12]. We binarized the language model using the KenLM tool [13]. In this set, we used the msd-bidirectional-fe model for lexical reordering. [14]

The Table 2 shows partial results of our experiments. We used shortcuts T (TED), E (EuroParl) and O (OpenSubtitles), if there is no additional suffix it means that test was baseline system trained on phrase-based model, suffix F (e.g. TF) means we used factored model, T refers to data that was true-cased and C means that a compound splitter was used. If suffix is I we used infinitive forms of all polish data and S suffix refers to changes in word order to meet SVO schema. In EuroParl experiments suffix L stands for bigger EN in-domain language model. H stands for highest score we obtained by combining methods and interpolating extra data. G suffix stands for tests on translation of our data by Google Translator.

**Table 2.** Experiment Results

|     | PL -> EN | | | | EN->PL | | | |
| --- | --- | --- | --- | --- | --- | --- | --- | --- |
|     | BLEU | NIST | MET | TER | BLEU | NIST | MET | TER |
| T   | 16,02 | 5,28 | 49,19 | 66,49 | 8,49 | 3,70 | 31,73 | 76,39 |
| TC  | 15,72 | 4,99 | 48,28 | 69,88 | 9,04 | 3,86 | 32,24 | 75,54 |
| TT  | 15,97 | 5,25 | 49,47 | 67,04 | 8,81 | 3,90 | 32,83 | 74,33 |
| TF  | 16,16 | 5,12 | 48,69 | 68,21 | 9,03 | 3,78 | 32,26 | 74,81 |
| TI  | 13,22 | 4,74 | 46,30 | 70,26 | 9,11 | 4,46 | 37,31 | 74,28 |
| TS  | 9,29 | 4,37 | 43,33 | 76,59 | 4,27 | 4,27 | 33,53 | 76,75 |
| TH  | 20,88 | 5,70 | 52,74 | 64,39 | 10,72 | 4,18 | 34,69 | 72,93 |
| TG  | 19,83 | 5,91 | 54,51 | 60,06 | 10,92 | 4,07 | 34,78 | 77,00 |
| E   | 73,18 | 11,79 | 87,65 | 22,03 | 67,71 | 11,07 | 80,37 | 25,69 |
| EL  | 80,60 | 12,44 | 91,07 | 12,44 | - | - | - | - |
| ELC | 80,68 | 12,46 | 90,91 | 16,78 | 67,69 | 11,06 | 80,43 | 25,68 |
| ELT | 78,09 | 12,41 | 90,75 | 17,09 | 64,50 | 10,99 | 79,85 | 26,28 |
| ELF | 80,42 | 12,44 | 90,52 | 17,24 | 69,02 | 11,15 | 81,83 | 24,79 |
| ELI | 70,45 | 11,49 | 86,21 | 23,54 | 70,73 | 11,44 | 83,44 | 22,50 |
| ELS | 61,51 | 10,65 | 81,75 | 31,71 | 49,69 | 9,38 | 69,05 | 40,51 |
| ELH | 82,48 | 12,63 | 91,17 | 15,73 | - | - | - | - |
| EG  | 32,87 | 7,57 | 68,45 | 50,57 | 22,95 | 6,01 | 60,75 | 46,33 |
| O   | 53,21 | 7,57 | 66,40 | 46,01 | 51,87 | 7,04 | 62,15 | 47,66 |
| OC  | 53,13 | 7,58 | 66,80 | 45,70 | - | - | - | - |
| OT  | 52,63 | 7,58 | 67,02 | 45,01 | 50,57 | 6,91 | 61,24 | 48,43 |
| OF  | 53,51 | 7,61 | 66,58 | 45,70 | 52,01 | 6,97 | 62,06 | 48,22 |
| OG  | 22,98 | 4,76 | 48,08 | 68,21 | 16,36 | 3,69 | 35,79 | 77,01 |

## 6 Discussion and Conclusions

We concluded that the results of the translations, in which the BLEU measure is greater than 70, can be considered as effective enough within the text domain. Such system already works in real time and can be connected into a pipeline with an automatic speech recognition and synthesis systems, which is our plan of future work.

Cleaning and converting of verbs to their infinitive forms improved EN-PL translation performance. However, this produced the opposite effect in PL- EN translation, perhaps due to reduction of the Polish vocabulary. Changing the word order to SVO is quite interesting. PL-EN translation scores degraded in this case, which we did not anticipate. On the other hand, some improvement could be seen in EN-PL translation. BLEU fell dramatically, and TER became slightly worse. NIST and METEOR showed better results than the baseline system. Hypothetically this is the result of each metric's evaluation method and that phrases were mixed in the SVO conversion phase. This phenomenon is worth further investigation.

Compound splitting proved to improve translation quality but mostly in PL-EN

translation. Factored training models also provide better translations but we gained improvement mostly in EN-PL experiments. Most likely reason is more complex Polish grammar. Truecasing did not act as anticipated, in most experiment scores were worse. We assume that data was already correctly cased.

In the future, there will be additional experiments performed with the use of extended language models. Tuning of training parameters for each set of data is required to be done separately (just like training higher order models). Training language model based on neural networks[1] also can be an interesting experiment.

Using other tools instead of GIZA, like Berkeley Aligner or Dyer's Fast Align or different phrase model (Hierarchical or Target Syntax), is also our plan for future work. We would also like to try out the factored training with Stemmed Word Alignment. Most probably using additional out of domain data and adapting it using for example Moore Levis Filtering could obtain further quality improvement.

# 7 Acknowledgements

This work is supported by the European Community from the European Social Fund within the Interkadra project UDA-POKL-04.01.01-00-014/10-00 and Eu-Bridge 7th FR EU project (grant agreement n°287658).

---

[1] www.fit.vutbr.cz/~imikolov/rnnlm